%% file: cvpr.tex
\documentclass[final]{cvpr}
\usepackage{float}
\usepackage{times}
\usepackage{epsfig}
\usepackage{graphicx}
\usepackage[font=small,skip=5pt]{caption}
\usepackage{subcaption}
\usepackage{amsmath}
\usepackage{amssymb}
\usepackage{bm}

\usepackage{pifont}
\usepackage{multirow}
\usepackage{makecell}
\usepackage[table,dvipsnames,svgnames,x11names]{xcolor}
\usepackage{booktabs}
\definecolor{dgreen}{rgb}{0.04,0.7,0.13}
\newcommand{\cmark}{{\color{red}\ding{51}}}

\newcolumntype{x}[1]{>{\centering\let\newline\\\arraybackslash\hspace{0pt}}p{#1}}
\usepackage{overpic} 

\usepackage[pagebackref=true,breaklinks=true,colorlinks,bookmarks=false]{hyperref}

\def\cvprPaperID{3521} %
\def\confYear{CVPR 2021}

\begin{document}
\title{Reconstructing 3D Human Pose by Watching Humans in the Mirror}

\author{Qi Fang\thanks{Equal Contribution. $^\dag$Corresponding author.}
\quad Qing Shuai$^{*}$
\quad Junting Dong
\quad Hujun Bao
\quad Xiaowei Zhou$^{\dag}$
\\[1.5mm]
State Key Lab of CAD\&CG, Zhejiang University 
}

\maketitle
\pagestyle{empty}
\thispagestyle{empty}

\begin{abstract}
In this paper, we introduce the new task of reconstructing 3D human pose from a single image in which we can see the person and the person's image through a mirror. Compared to general scenarios of 3D pose estimation from a single view, the mirror reflection provides an additional view for resolving the depth ambiguity.
We develop an optimization-based approach that exploits mirror symmetry constraints for accurate 3D pose reconstruction. We also provide a method to estimate the surface normal of the mirror from vanishing points in the single image. 
To validate the proposed approach, we collect a large-scale dataset named Mirrored-Human, which covers a large variety of human subjects, poses and backgrounds. The experiments demonstrate that, when trained on Mirrored-Human with our reconstructed 3D poses as pseudo ground-truth, the accuracy and generalizability of existing single-view 3D pose estimators can be largely improved. The code and dataset are available at \url{https://zju3dv.github.io/Mirrored-Human/}.

\end{abstract}

\section{Introduction}
\input{src/01_introduction}

\section{Related work}
\input{src/02_relatedwork}

\section{Methods}

\input{src/03_method}

\input{src/04-1_evaulation}

\input{src/05-1_learning}

\input{src/06_conclusion}

\newpage
{\small
\bibliographystyle{ieee_fullname}
\bibliography{cvpr}
}

\end{document}

%% file: src/01_introduction.tex
3D human pose estimation has ubiquitous applications in sport analysis, human-computer interaction, and fitness and dance teaching. While there has been remarkable progress in 3D pose estimation from a monocular image or video~\cite{hmrKanazawa17, Moon_2020_ECCV_I2L-MeshNet, kolotouros2019spin, kocabas2019vibe, xiang2019monocular}, inevitable challenges such as the depth ambiguity and the self-occlusion are still unsolved.

\begin{figure}
     \centering
     \begin{subfigure}[h]{0.23\textwidth}
         \centering
         \includegraphics[width=\textwidth]{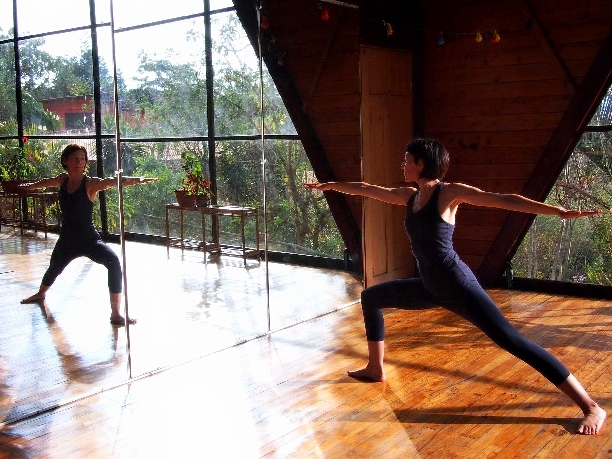}
         \caption*{Input image}
     \end{subfigure}
     \begin{subfigure}[h]{0.23\textwidth}
         \centering
         \includegraphics[width=\textwidth]{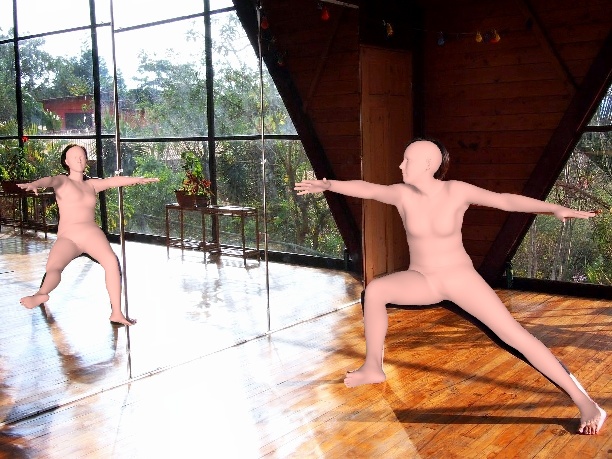}
         \caption*{SMPLify-X~\cite{SMPL-X:2019}}
     \end{subfigure}
     \vspace*{0.2cm}
     \begin{subfigure}[h]{0.45\textwidth}
         \centering
         \includegraphics[width=0.98\linewidth, trim=25 50 25 50]{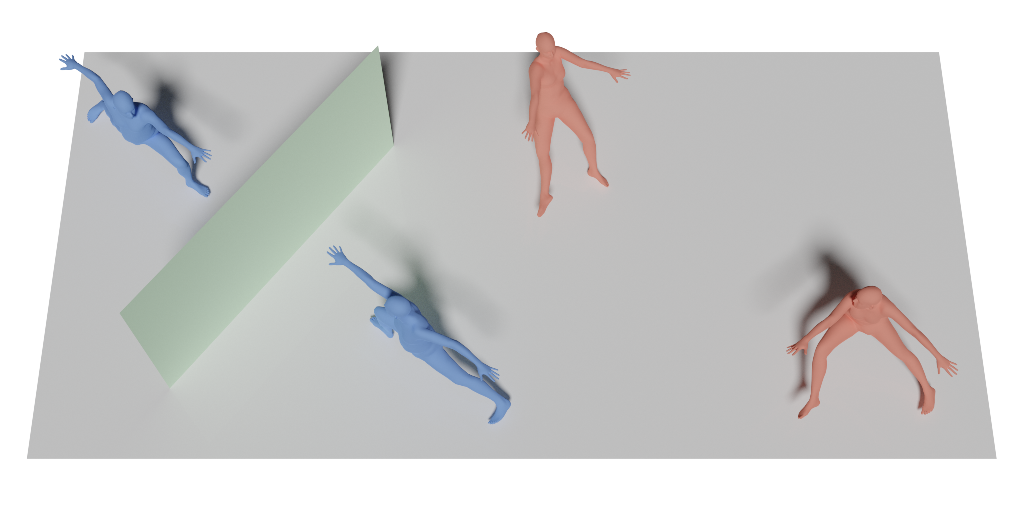}
         \caption*{3D visualization of our (left) and SMPLify-X (right) results}
     \end{subfigure}
     \vspace*{-0.2cm}
     \caption{While the state-of-the-art single-view 3D pose estimator~\cite{SMPL-X:2019} yields a small reprojection error, the recovered 3D poses may be erroneous due to the depth ambiguity. We make use of the mirror in the image to resolve the ambiguity and reconstruct more accurate human pose as well as the mirror geometry.}
     \vspace*{-0.5cm}
    \label{fig:demo1}
\end{figure}

In many scenes like dancing rooms and gyms, people are often in front of a mirror. In this case, we are able to see the person and his/her mirror image simultaneously. The mirror image actually provides an additional virtual view of the person, which can resolve the single-view depth ambiguity if the mirror is properly placed. Moreover, unseen part of the person can also be observed from the mirror image, so that the occlusion problem can be alleviated.

In this paper, we investigate the feasibility of leveraging such mirror images to improve the accuracy of 3D human pose estimation. We develop an optimization-based framework with mirror symmetry constraints that are applicable without knowing the mirror geometry and camera parameters. We also provide a method to utilize the properties of vanishing points to recover the mirror normal along with the camera parameters, so that an additional mirror normal constraint can be imposed to further improve the human pose estimation accuracy. The effectiveness of our framework is validated on a new dataset for this new task with 3D pose ground-truth provided by a multi-view camera system.

An important application of the proposed approach is to generate pseudo ground-truth annotations to train existing 3D pose estimators. To this end, we collect a large-scale set of Internet images that contain people and mirrors and generate 3D pose annotations with the proposed optimization method. The dataset is named Mirrored-Human.  
Compared with existing 3D human pose datasets~\cite{h36m_pami,mono-3dhp2017,vonMarcard2018} that are captured with very few subjects and background scenes, Mirrored-Human has a significantly larger diversity in human poses, appearances and backgrounds, as shown in Fig.~\ref{fig:dataset}. The experiments show that, by combining Mirrored-Human with existing datasets as training data, both accuracy and generalizability of existing 3D pose estimation methods can be significantly improved for both single-person and multi-person cases.   

In summary, we make the following contributions:
\begin{itemize}
    \item We introduce a new task of reconstructing  human pose from a single image in which we can see the person and the person's mirror image. 
    \item We develop a novel optimization-based framework with mirror symmetry constraints to solve this new task, as well as a method to recover mirror geometry from a single image.
    \item We collect a large-scale dataset named Mirrored-Human from the Internet, provide our reconstructed 3D poses as pseudo ground-truth, and show that training on this new dataset can improve the performance of existing 3D human pose estimators. 
\end{itemize}

%% file: src/02_relatedwork.tex
\begin{figure*}[t]
	\centering
	\includegraphics[width=1\linewidth,trim={0cm 0cm 0cm 0.5cm},clip]{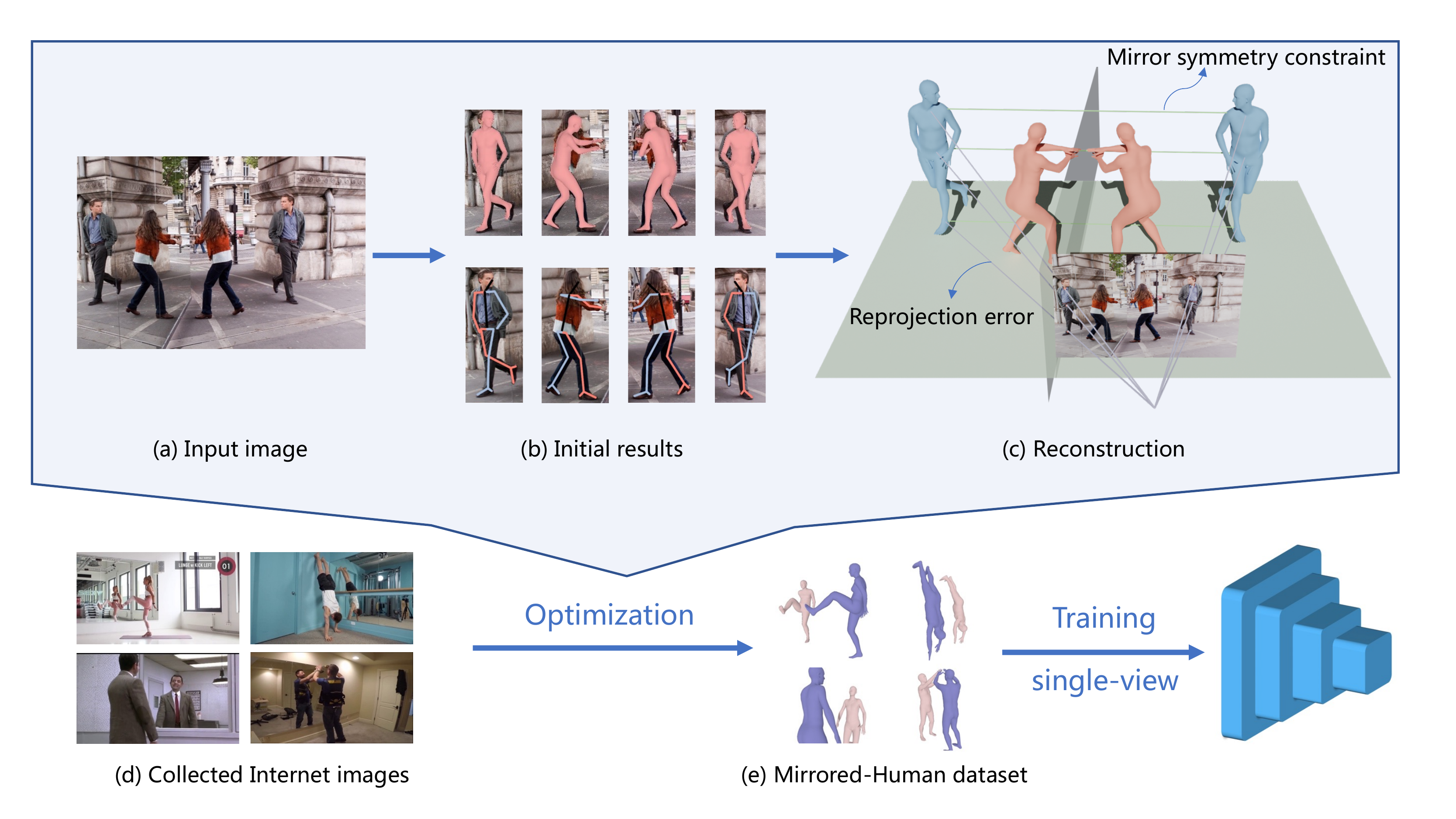}
	\vspace{-0.8cm}
	\caption{\textbf{Overview of our approach.} Given the input image (a), we first estimate the 2D keypoints and SMPL parameters as the initialization (b). Then, we minimize the reprojection error with mirror symmetry constraints for reconstruction (c). We collect a considerable number of Internet images (d) and build a dataset named Mirrored-Human (e) with pseudo ground-truth generated by our framework. The dataset can be used for the training of single-view methods.}
	\label{fig:method}
\end{figure*}

\subsection{3D human pose}
Benefiting from neural networks, the task of monocular 3D human pose estimation has made considerable progress. The skeleton-based approaches either lift the 2D pose to 3D~\cite{martinez_2017_3dbaseline, chen20173d, wandt2019repnet}, or adopt an end-to-end manner to regress the 3D pose directly from the image~\cite{sun2017compositional, sun2018integral, zhou2017towards, pavlakos2018ordinal}. The model-based approaches estimate the pose and shape simultaneously with parametric models~\cite{anguelov2005scape, SMPL:2015, STAR:2020, xu2020ghum}. Approaches along this line can be divided into two categories. Optimization-based methods fit the model using 2D evidence and some human body priors~\cite{Bogo:ECCV:2016, Lassner:UP:2017, SMPL-X:2019, dong2020motion}. Regression-based methods directly regress the model parameters from the image~\cite{hmrKanazawa17, omran2018nbf, xu2019denserac, humanMotionKanazawa19, zeng20203d}. Kolotouros~\etal~\cite{kolotouros2019spin} incorporate their advantages and propose a self-improving framework. To relax the heavy reliance on the model's parameter space, model-free approaches directly regress the 3D locations of the mesh vertices ~\cite{kolotouros2019cmr, Moon_2020_ECCV_I2L-MeshNet,Choi_2020_ECCV_Pose2Mesh}. For multi-person cases, previous works focus on how to extend single-person frameworks to multi-person ones~\cite{rogez2019lcr, mehta2019xnect} or model the interaction between people~\cite{zanfir2018monocular, fieraru3d}. Some recent works~\cite{Moon_2019_ICCV_3DMPPE, zhen2020smap, li2020hmor, lin2020hdnet, fabbri2020compressed} explore the representation of the absolute depth in the camera coordinate system.

In monocular settings, the depth ambiguity is inevitable. One solution is to utilize the additional supervision signal. Pavlakos \etal \cite{pavlakos2018ordinal} use the ordinal depths of human joints to weakly supervise the network. Kanazawa \etal \cite{hmrKanazawa17} train a discriminator network to judge if the estimated pose is reasonable. Some other works use temporal constraints \cite{kocabas2019vibe, tripathi2020posenet3d, peng2021neural} or geometric self-consistency \cite{chen2019unsupervised}.
Another line of work resolves the ambiguity with scene layouts. Hassan \etal \cite{PROX:2019} exploit static 3D scene structures with the inter-penetration and the contact constraints. Others \cite{mehta2019xnect, zanfir2018monocular} integrate the ground plane information to recover the 3D location and pose. We use the additional view provided by the mirror reflection to resolve the depth ambiguity.

Learning-based methods are inseparable from the training data. Existing widely-used 3D datasets such as Human3.6M~\cite{h36m_pami}, Human Eva~\cite{Sigal:IJCV:10b}, 
MPI-INF-3DHP~\cite{mono-3dhp2017} and Panoptic Studio~\cite{Joo_2017_TPAMI} are built with motion capture systems and thus have limited appearance and pose diversity. 3DPW~\cite{vonMarcard2018} is a recent dataset consisting of ordinary activities. The combination of a camera and several IMUs attached to the human body provides accurate 3D poses, but the dataset still lacks diversity. Kanazawa~\etal~\cite{humanMotionKanazawa19} contribute some Internet datasets which however lack 3D annotations. Arnab~\etal~\cite{Arnab_CVPR_2019} propose a bundle-adjustment-based algorithm, based upon which they generate 3D annotations for Internet data. However, this method still suffers from the depth ambiguity derived from monocular videos. To summarize, a \textit{large-scale} \textit{Internet} dataset with \textit{3D annotations} is still missing. We collect numerous Internet images containing mirrors and people with a large diversity in appearances and poses, and build a dataset using our reconstructed 3D poses as pseudo ground-truth, to address this problem to some extent.

\subsection{Reconstruction with mirrors}
Earlier years have witnessed some researches on the mirror geometry of a catadioptric system (mirrors + lenses), especially for reconstruction and extrinsic camera calibration~\cite{takahashi2012, rodrigues2010}. For 3D reconstruction that is more relevant to our task, some works use the configuration of two planar mirrors to capture stereo images~\cite{nene1998, gluckman2001catadioptric, tahara2015interference, lanman2009surround} or produce four virtual views by assuming there is a one-time interreflection between two mirrors~\cite{ying2012self}. They calibrate the mirror based on the image correspondences or silhouettes. Nguyen~\etal~\cite{nguyen20183d} use a depth sensor and at least two mirrors, trying to remove depth distortions. Akay~\etal~\cite{akay2014} reconstruct 3D object with a mirror and RGBD cameras. Hu~\etal~\cite{Hu2005MultipleView3R} only use one mirror and one RGB camera, but they need to label the object and mirror region. To the best of our knowledge, the task of exploiting mirror symmetry to recover 3D human pose and shape from an Internet image has not been discussed in literature. 

Furthermore, not only does the symmetry property exist in scenes with mirrors, but it also appears on symmetrical objects, which has been explored to reconstruct faces, cars, etc. Some works ~\cite{sinha2012detecting, jiang2009symmetric} explicitly detect the plane of symmetry with 2D correspondences and others~\cite{Wu_2020_CVPR} implicitly use the symmetry prior.

%% file: src/03_method.tex
Fig.~\ref{fig:method} presents the pipeline of our framework. Given an image that contains a person and a mirror, our goal is to recover the human mesh considering the mirror geometry. The key insight is that the person and his/her mirror image can be treated as two people, and we reconstruct them together with the mirror symmetry constraints. This section will be organized as follows. First, the formulation of single-person mesh recovery is introduced (Sec.~\ref{sec:spmr}). Then the mirror symmetry constraints that relate the two people will be elaborated (Sec.~\ref{sec:mi_geo}). Finally, the objective functions and the whole optimization are described (Sec.~\ref{sec:opt}).

\subsection{Human mesh recovery with SMPL model}
\label{sec:spmr}

We adopt the SMPL model~\cite{SMPL:2015} as our human representation. The SMPL model is a differentiable function $\bm M(\bm \theta, \bm \beta) \in \mathbb R^{3\times N_v}$ mapping the pose parameters $\bm \theta \in \mathbb R^{72}$ and the shape parameters $\bm{\beta} \in \mathbb R^{10}$ to a triangulated mesh with $N_v = 6890$ vertices. The 3D body joints $\bm J(\bm\theta, \bm\beta)$ of the model can be defined as a linear combination of the mesh vertices. Hence for $N_j$ joints, we defined the body joints $\bm J(\bm\theta, \bm\beta) \in \mathbb{R}^{3\times N_j} = \mathcal{J}(\bm M(\bm\theta, \bm\beta))$, where $\mathcal{J}$ is a pre-trained linear regressor. Let $\bm R \in SO(3)$ and $\bm T \in \mathbb R^3$ denote the global rotation and translation, respectively.

Given an image and the detected 2D bounding boxes, the 2D human keypoints $\bm W$ can be estimated with the cropped regions. The objective function for human mesh recovery generally consists of a reprojection term $L_{2d}$ and a prior term $L_p$ with respect to variables $\bm \theta$, $\bm \beta$, $\bm R$ and $\bm T$.

The reprojection term penalizes the weighted 2D distance between the estimated 2D keypoints $\bm{W}$ with the confidence $c$, and the corresponding projected SMPL joints:
\begin{equation}
    L_{2d} = \sum_i c_i\rho(\bm{W}_i - \Pi_K(\bm{R}\bm{J}(\bm\theta, \bm\beta)_i + \bm{T})),
\end{equation}
where $\Pi_K$ is the projection from 3D to 2D through the intrinsic parameter $K$. $\rho$ denotes the Geman-McClure robust error function for suppressing noisy detections. 

The human body priors are used to encourage realistic 3D human mesh results. Since the pose and shape parameters ($ \bm{\Tilde{\theta}}, \bm{\Tilde{\beta}}$) estimated by a neural network can be viewed as learned prior, the final results are supposed to be close to them:
\begin{equation}
    L_{p} = ||\bm{\theta} - \bm{\Tilde{\theta}}||_2^2  + \lambda_{\beta}|| \bm{\beta} - \bm{\Tilde{\beta}}||_2^2,
\end{equation}
where $\lambda_{\beta}$ is a weight.

\subsection{Mirror-induced constraints}
\label{sec:mi_geo}
\begin{figure}[t]
\centering
\includegraphics[trim=3cm 18.5cm 10cm 3.5cm, width=0.8\linewidth,clip]{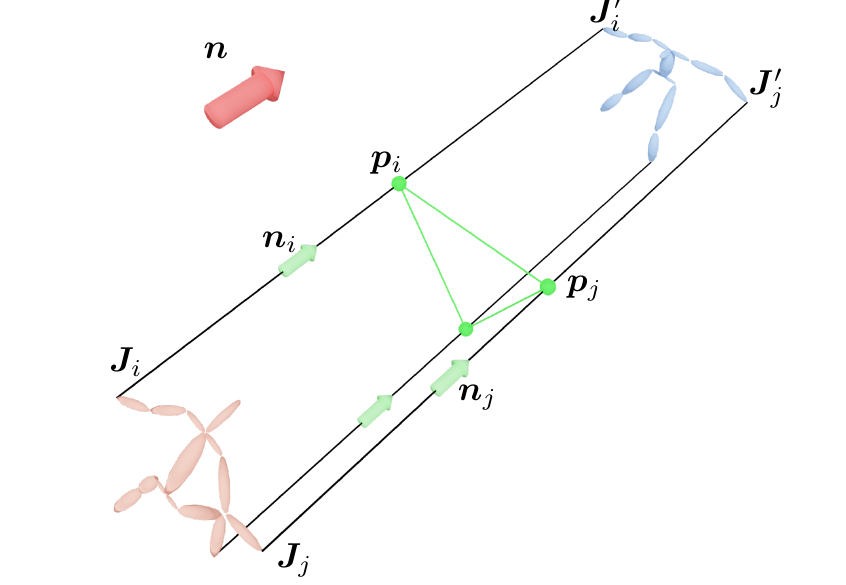}
\caption{\textbf{An illustration of mirror-induced constraints.} The line segment connecting the joint $\bm{J}_i$ and its mirrored joint $\bm{J}_i'$ has the direction $\bm n_i$ and the middle point $\bm p_i$. Theoretically, $\bm{n}_i // \bm{n}_j$, and $\bm{n}_i \perp \overline{\bm{p}_i\bm{p}_j}$. If the mirror normal $\bm{n}$ (red arrow) is known, $\bm{n} // \bm{n}_i$ and $\bm{n} // \bm{n}_j$ should be satisfied as well.}
\label{fig:mirrorsym}
\end{figure}

If there is a mirror in the image, the relation between the person and the mirrored person can be used to enhance the reconstruction performance. This relation is a simple reflection transformation if the mirror geometry is known, which however is impracticable for an arbitrary image from the Internet. To tackle this problem and take advantage of the characteristic of the mirror, the following mirror-induced constraints are introduced, as illustrated in Fig.~\ref{fig:mirrorsym}. Note that all symbols with the superscript prime refer to variables related to the mirrored person unless specifically mentioned.

\paragraph{Mirror symmetry constraints:}
Since the adopted human representation disentangles the orientation $\bm R$, pose parameters $\bm \theta$ and shape parameters $\bm \beta$, $\bm \beta$ can be shared by the person and the mirrored person, and $\bm \theta$ is related to $\bm \theta'$ by a simple reflection operation as follows:
\begin{equation}
\label{eq:param}
    \bm{\beta}' = \bm{\beta}, ~\bm{\theta}' = \mathcal{S}(\bm \theta),
\end{equation}
where $\mathcal{S}(\cdot)$ denotes the reflection operation on axis angles. 

As Eq.~\ref{eq:param} does not take $\bm R$ and $\bm{T}$ into consideration, the constraint on 3D keypoints can be imposed to estimate the human orientation and position better. We abbreviate the global coordinates of the $i$-th joint $\bm{R}\bm{J}(\bm\theta, \bm\beta)_i + \bm{T}$ as $\bm{J}_i$. Given a pair of body joints $i, j$, we denote the direction of the line segment $\overline{\bm{J}_i\bm{J}_i'}$, $\overline{\bm{J}_j\bm{J}_j'}$ as $\bm{n}_i$, $\bm{n}_j$ and the middle point of them as $\bm p_i$, $\bm p_j$, respectively. Ideally, $\bm n_i$ should be parallel to $\bm n_j$ and $\bm p_i, \bm{p}_j$ are supposed to be on the mirror plane. Despite the fact that the mirror geometry is unknown, it needs to be satisfied that $\bm{n}_i$ is perpendicular to the line  $\overline{\bm{p}_i\bm{p}_j}$. So for 
any pair of joints, we minimize the sum of the L2 norm of the cross product between $\bm{n}_i$ and $\bm{n}_j$, and the inner product between $\bm{n}_i$ and $\bm{p}_j - \bm{p}_i$: 
\begin{equation}\label{eq:mirrorsym}
    L_{s} = \sum_{(i, j)}(||\bm{n}_i \times \bm{n}_j||_2 + || \bm{n}_i\cdot (\bm{p}_j - \bm{p}_i) ||_2).
\end{equation}

\paragraph{Mirror normal constraint:}
A mirror can be represented as a plane, parameterized as its normal and position. If its normal $\bm{n}$ is known, the geometric properties of the mirror can thus be utilized explicitly by constraining $\bm n_i$ and $\bm n$ to be parallel with the following loss function:
\begin{equation}\label{eq:mirrorgt}
    L_{n} =  \sum_i||\bm{n} \times \bm{n}_i||_2. 
\end{equation}
\vspace{-0.5cm}
\begin{figure}[t]
	\centering
	\includegraphics[width=1\linewidth,trim={8cm 8cm 8cm 7.5cm},clip]{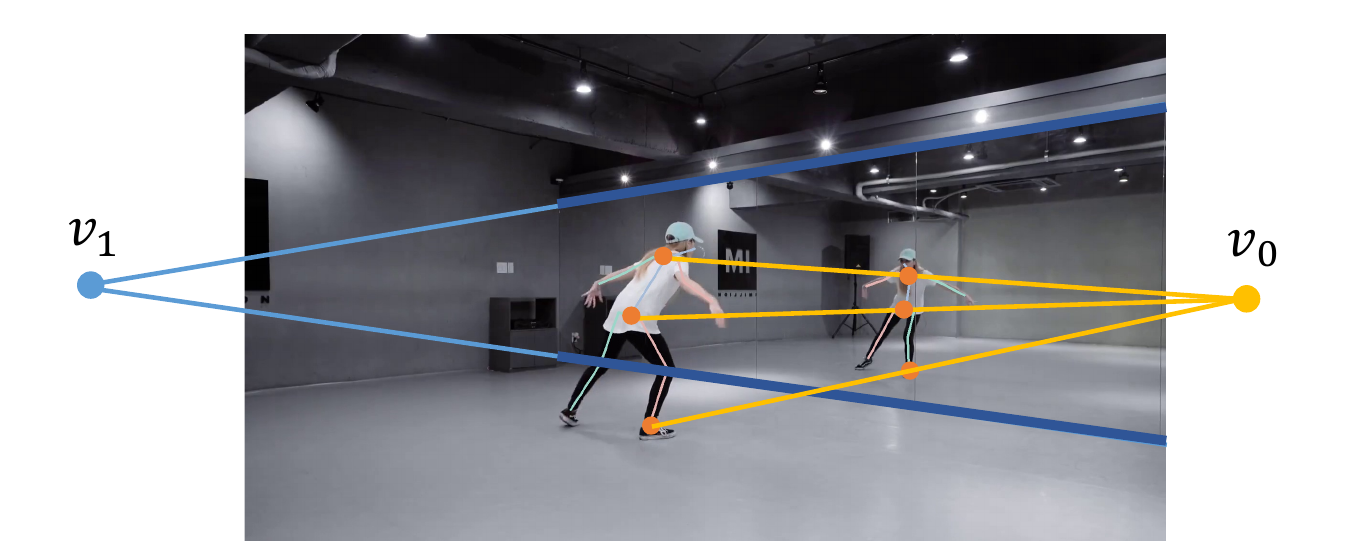}
	\includegraphics[width=\linewidth, trim={2.5cm 2cm 1cm 3cm},clip]{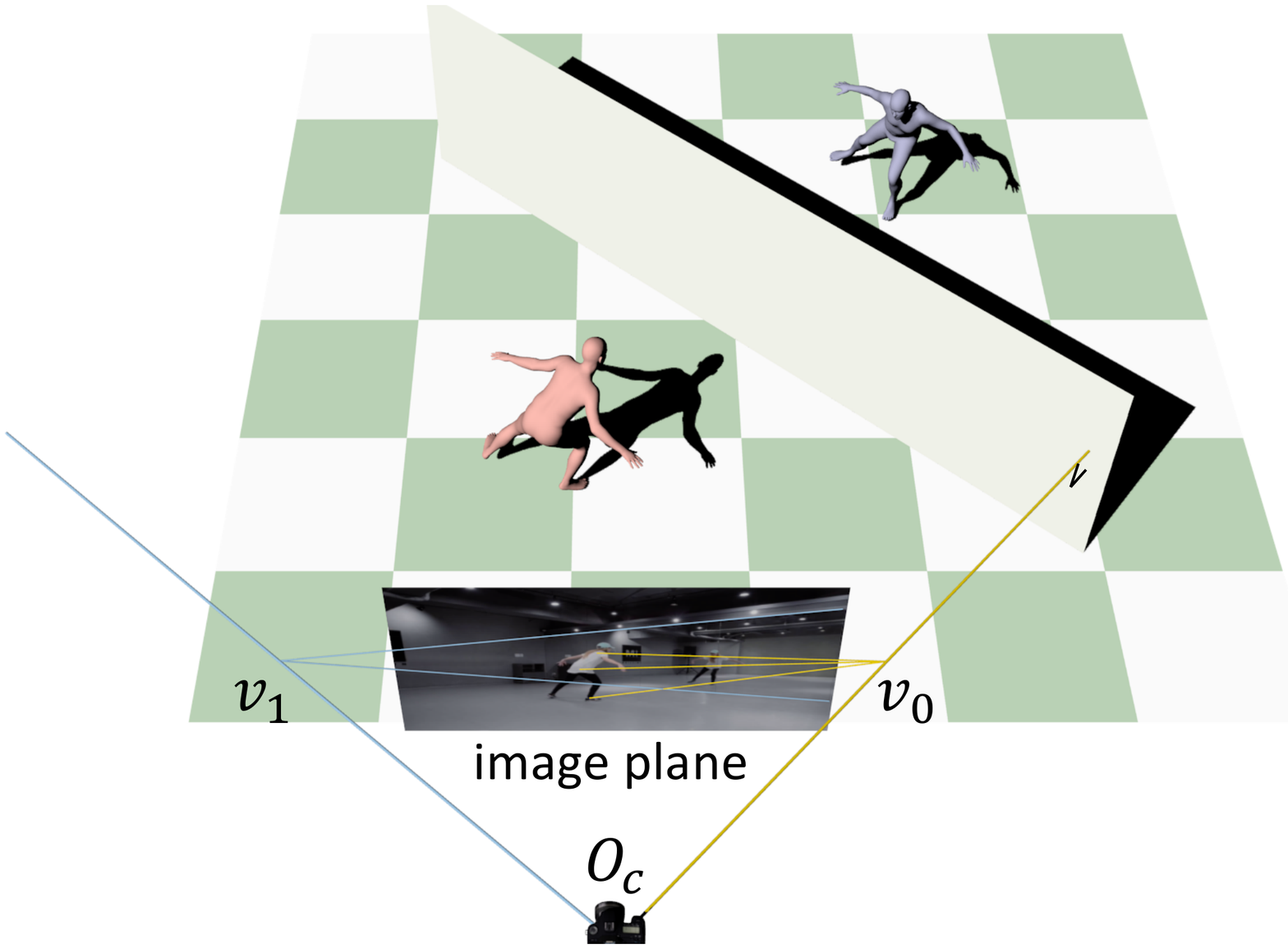}
	\vspace{-0.8cm}
	\caption{\textbf{Vanishing points in an image containing a person and a mirror.} In most cases at least two vanishing points can be found, where $\bm v_0$ comes from 2D human keypoints, and $\bm v_1$ comes from the annotated mirror edges. $O_c$ denotes the camera center. Note that $\overline{O_c v_0} // \bm n$ and $\overline{O_c v_1} \perp \bm n$, where $\bm n$ is the mirror normal.
	}
	\label{fig:vp}
\end{figure}
\paragraph{Mirror normal estimation:}
Though the mirror normal is not directly available, the vanishing points can be used to estimate it. The vanishing point of lines with direction $\bm n$ in 3D space is the intersection $\bm v$ of the image plane with a ray through the camera center with direction $\bm n$~\cite{hartley2003multiple}:
\begin{equation}
\bm v=K \bm n,
\label{eq:vanish}
\end{equation}
where the vanishing point $\bm v\in \mathbb{R}^3$ is in the form of homogeneous coordinates and $K$ is the camera intrinsic matrix. 

Eq.~\ref{eq:vanish} reveals that obtaining the mirror normal $\bm n$ requires both $K$ and $\bm v$. As the parallel lines connecting points on the real object and corresponding points on the mirrored object are perpendicular to the mirror, the vanishing point $\bm v$ with this direction can be estimated through their 2D positions.
To get such correspondences, some previous works require additional inputs such as masks \cite{Hu2005MultipleView3R}, which is infeasible for images from the Internet.
Fortunately, since 2D human keypoints provide robust semantic correspondences, \eg the left ankle of the real person and the right ankle of the mirrored person, this vanishing point can be acquired naturally and automatically in our setting ($\bm v_0$ in Fig.~\ref{fig:vp}).

Note that if the intrinsic matrix $K$ is provided, the mirror normal can thus be solved easily through $\bm n=K^{-1} \bm v_0$, otherwise $K$ should be calibrated from a single image if possible. From the projective geometry~\cite{hartley2003multiple}, we know that it is possible to calibrate the camera intrinsic parameters from a single image. Suppose the camera has zero skew and square pixels. The intrinsic matrix $K$ can be computed via three orthogonal vanishing points. Additionally, if the principal point is assumed to be in the image center (only the focal length is unknown), $K$ can be computed via only two orthogonal vanishing points. Please refer to the supplementary material for more details. 

As we have stated, one vanishing point $\bm v_0$ has been acquired based on reliable 2D human keypoints. Different from the general scene where finding orthogonal relations may be difficult, our setting contains richer information. Fig.~\ref{fig:vp} shows that if we annotate the mirror edges, at least one vanishing point $\bm v_1$ orthogonal to $\bm v_0$ can be obtained. With these vanishing points, the calibration can be performed. Note that images from the same video share the same intrinsic matrix $K$, thus the annotation process is not laborious.

The mirror normal constraint is optional, which depends on how easy it is to find mirror edges. In the experiment, we will show that our method can still achieve satisfactory performance without the mirror normal constraint.

\subsection{Objective function and optimization}
\label{sec:opt}
Combining all discussed above, the final objective function to optimize can be written as:
\begin{equation}
\label{eq:loss0}
\begin{split}
    \min_{\substack{\Theta, \Theta'}}~L_{2d}+L_{2d}' + \lambda_p (L_{p}&+L_{p}') + \lambda_s L_s + \lambda_n L_n \\
     s.t.~~ \bm{\beta}' = \bm{\beta}&, ~\bm{\theta}' = \mathcal{S}(\bm \theta),
\end{split}
\end{equation}
where $\Theta=\{\bm\theta, \bm\beta, \bm R, \bm T\}$ and $\Theta'=\{\bm\theta', \bm\beta', \bm R', \bm T'\}$. $L_{2d}'$ and $L_p'$ refer to the reprojection term and the prior term of the mirrored person, respectively. $\lambda_p$, $\lambda_s$ and $\lambda_n$ are weights. $\lambda_n$ is set to zero whenever the mirror normal is unavailable. If there are two or more people, the optimization can be done for each subject separately.

We optimize Eq.~\ref{eq:loss0} with respect to all parameters using L-BFGS and PyTorch. An off-the-shelf model~\cite{kolotouros2019spin} is adopted to generate the initial estimation. Given the 2D keypoints~\cite{sun2019deep, cao2017realtime}, $\bm R$ and $\bm T$ are further optimized by aligning the initial SMPL model to the 2D keypoints. To improve the robustness of the initialization, we select the person with smaller reprojection error and apply the selected pose parameter to the other person after a reflection operation.

%% file: src/04-1_evaulation.tex
\section{Evaluation}
\begin{figure}[t]
	\centering
	\includegraphics[width=1.0\linewidth,trim={0cm 0cm 0cm 0cm},clip]{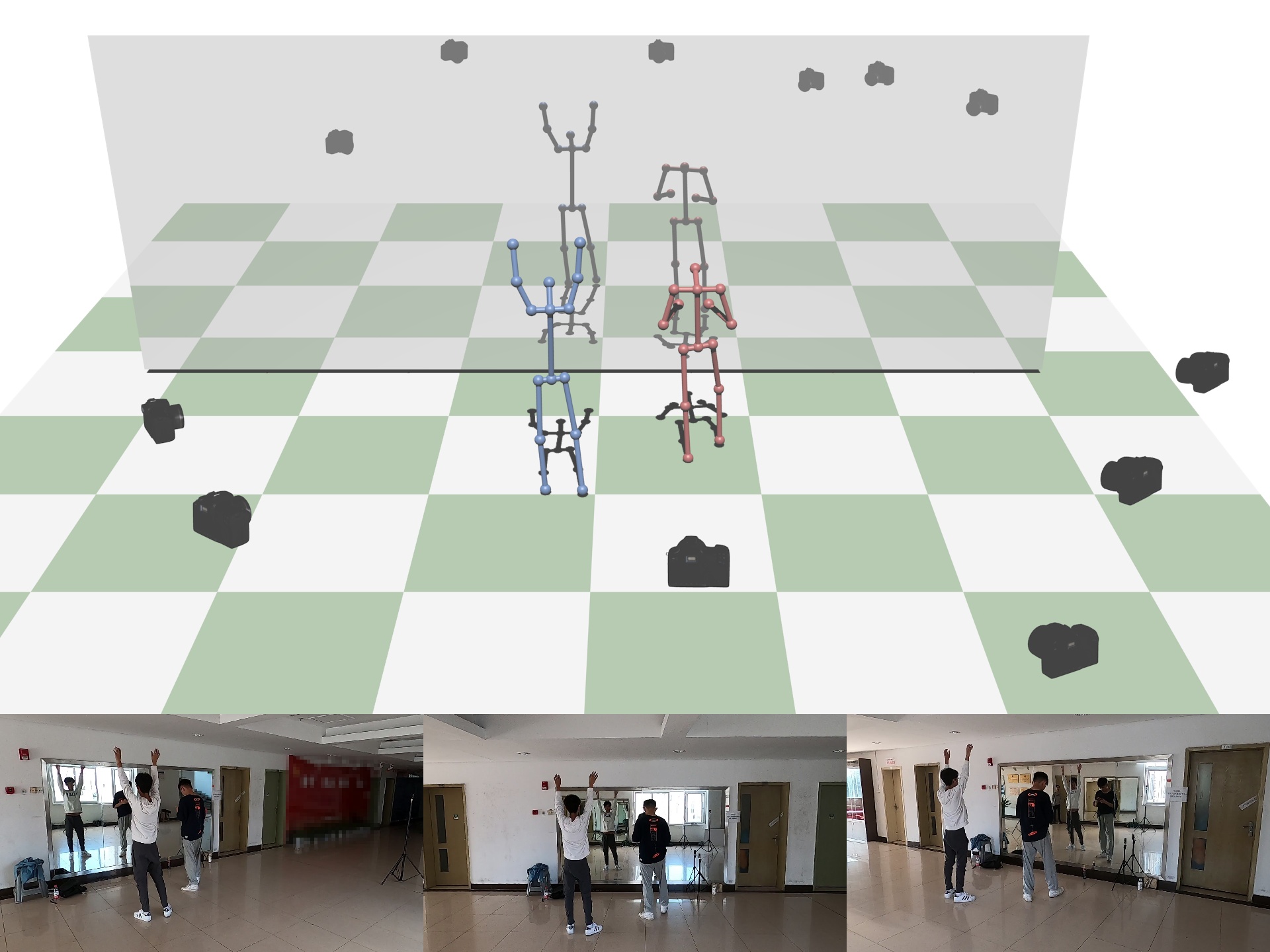}
	\vspace{0.2mm}
	\caption{
	\textbf{The configuration of our evaluation set.} The poses of all subjects are captured with six cameras. The camera extrinsic paramters, camera intrinsic parameters, and mirror geometry are calibrated.
	}
	\label{fig:mv_mirror}
\end{figure}

\subsection{Dataset for evaluation}
Since no dataset exists for our task which contains both mirrors and labeled 3D human keypoints, we collect a new dataset with six calibrated HD cameras, as shown in Fig.~\ref{fig:mv_mirror}. All videos are recorded with a resolution of 1920$\times$1080 pixels at 30 fps. 
Each person is performing various actions in front of a large mirror. A calibration board is placed on the mirror, thus the mirror geometry can be easily determined, which can also be used to evaluate the estimated mirror geometry. Note that the ground-truth of 3D keypoints is generated from all views.
2D bounding boxes, 2D keypoints, and the correspondences between multiple people are annotated manually.

\subsection{Reconstruction evaluation and ablation study}
\label{sec:mv-mirror}

The 3D keypoints generated by this multi-view system are used as ground-truth to evaluate the reconstruction accuracy. Metrics include MPJPE, PA-MPJPE, and MRPE. MPJPE is the distance (mm) between predicted and ground-truth 3D keypoints after root alignment. PA-MPJPE is calculated with further Procrustes alignment. MRPE is defined as the distance (mm) between the predicted and the ground-truth root joint, which evaluates the absolute root position accuracy.
\input{tables/quant_ablation}

\begin{figure}[t]
	\centering
	\includegraphics[width=0.8\linewidth, trim=10 0 50 50,clip]{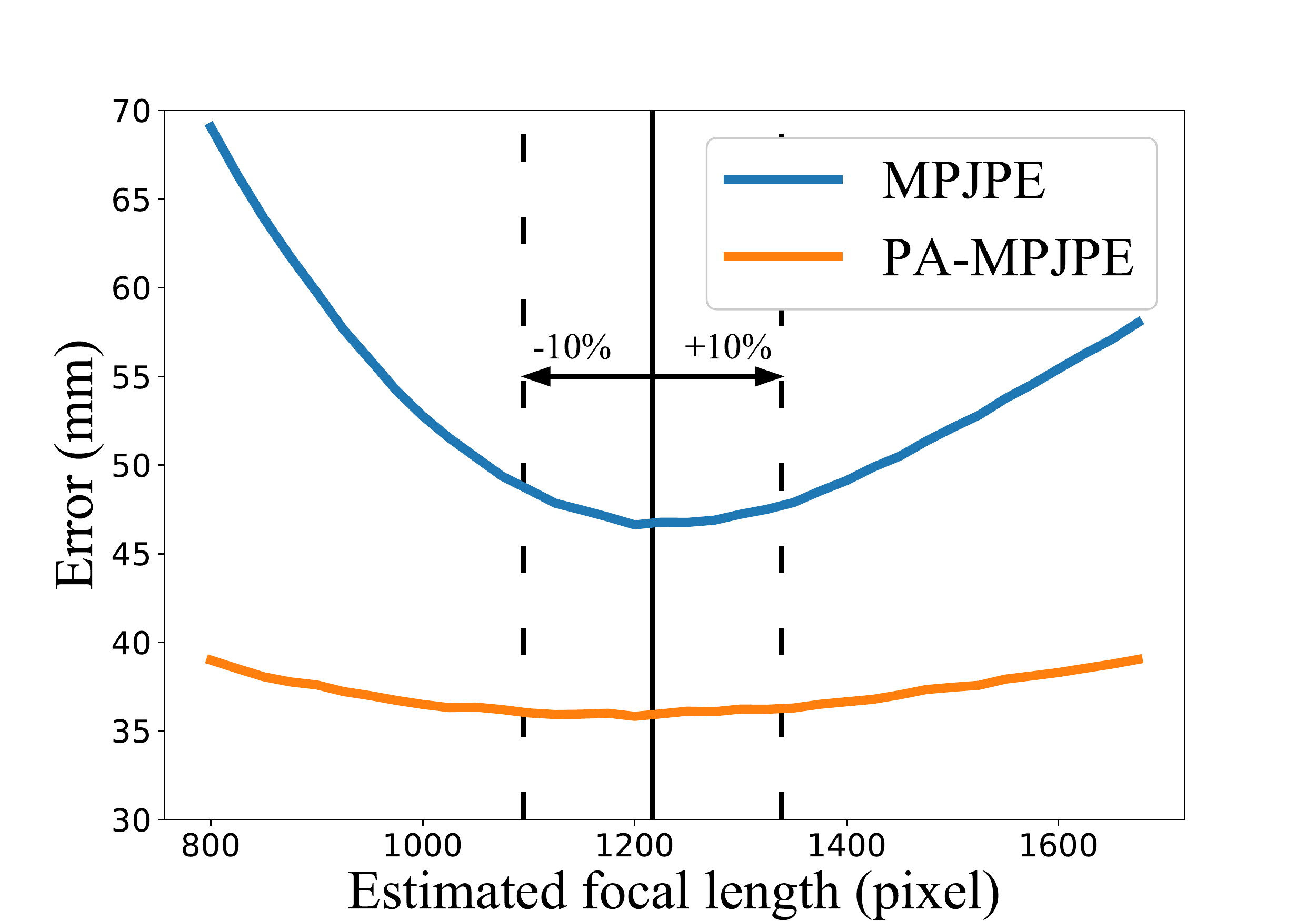}
	\caption{\textbf{Sensitivity of pose error to focal length estimation.} The vertical black line indicates the ground-truth focal length.}
	\vspace{-1mm}
	\label{fig:focal}
\end{figure}
\paragraph{Pose Reconstruction:} 
Two previous methods are compared here. SMPLify-X~\cite{SMPL-X:2019} is the state-of-the-art optimization-based method and SPIN~\cite{kolotouros2019spin} is the state-of-the-art regression-based method. `Baseline' combines the two methods by using \cite{kolotouros2019spin} as initialization and \cite{SMPL-X:2019} as optimization. In Table~\ref{tab:ablation}, comparing `Ours (full)' with first three rows, the result shows that our approach outperforms previous methods by a large margin. 

The ablation study is performed to show the effect of our mirror-induced constraints. It can be observed in Table~\ref{tab:ablation} that without the mirror normal constraint (`Ours (w/o $L_{n}$)'), our method can still perform well, indicating our applicability in the case where vanishing points are hard to acquire. If $L_{s}$ is also discarded (`Ours (w/o $L_{s}$, $L_{n}$)'), MPJPE will degrade severely while PA-MPJPE has a relatively slight change, reflecting that the mirror symmetry constraint can adjust the human orientation effectively. `Ours (w/o $L_{s}$, $L_{n}$)' is better than the baseline since $\bm \theta$ and $\bm \beta$ are shared. For MRPE, more constraints will bring the better improvement and the position error of our full model is less than 10cm. 

We also measure the accuracy of mirror normal estimation. As we have stated, the ground-truth mirror normal comes from calibration. We use the average angle between the ground-truth and the estimated normal. The angle is 1.9$^\circ$ with the ground-truth focal length, and 4.1$^\circ$ with the estimated focal length, both of which are quite small.

\paragraph{Focal length estimation:} To evaluate the accuracy of focal length estimation, we simulate the case in which only the focal length is unknown and two orthogonal vanishing points are used. One vanishing point is computed by the lines parallel to the mirror and the other is from two ways: 1) if we label two lines that are perpendicular to the mirror from the scene, the error of focal length is 4.9\% of the true value, and 2) if we make use of 2D human keypoints, the error reduces to 3.6\%. It means that the 2D human keypoints provide reliable correspondences that are helpful to estimate the focal length.

To evaluate the influence of the predicted focal length on pose estimation, we modify the focal length to different values and calculate the reconstruction accuracy. As shown in Fig.~\ref{fig:focal}, when the focal length deviates from the ground-truth by less than 10\%, the change of the reconstruction error is less than 5\%, showing the stability of our algorithm.
\begin{figure}[t]
	\centering
	\includegraphics[width=1\linewidth]{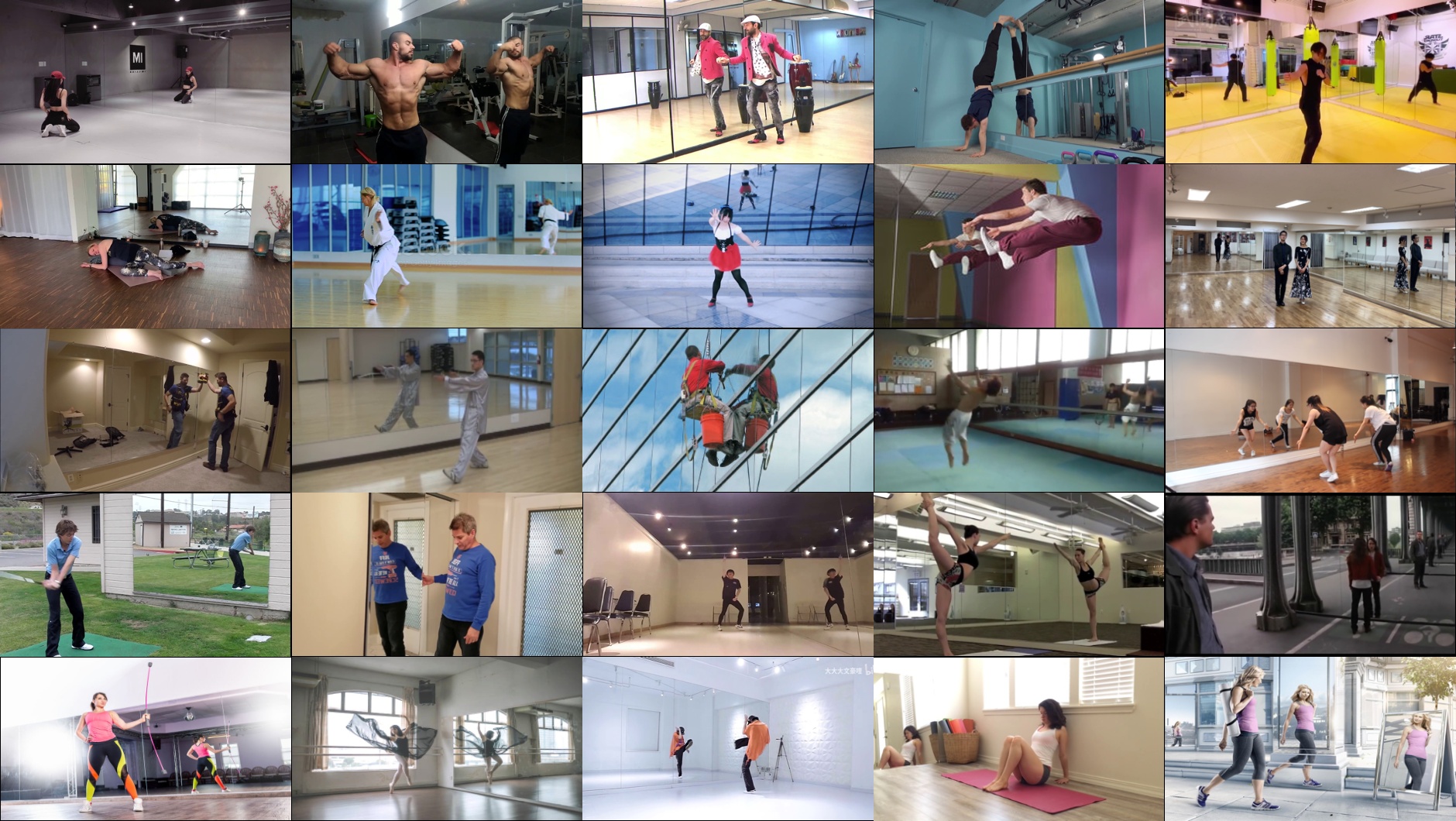}
	\caption{\textbf{Samples in our Mirrored-Human dataset.} From left to right, each column shows the variety of actions, appearances, viewpoints, poses and multiple mirrors/people, respectively. }
	\label{fig:dataset}
\end{figure}
\input{tables/dataset}

\subsection{Qualitative comparison}
Some representative results are visualized in Fig.~\ref{fig:results}, showing that our method can reconstruct accurate 3D human meshes as well as the mirror geometry. 
Compared with the monocular human mesh recovery algorithms~\cite{SMPL-X:2019, kolotouros2019spin}, our method produces much more consistent positions and orientations  (Fig.~\ref{fig:results}) and more accurate poses (Fig.~\ref{fig:comparisonpose}). More qualitative results can be found in the supplementary material.

%% file: tables/quant_ablation.tex
\begin{table}[t]
	\begin{center}
	\resizebox{\columnwidth}{!}{
	\begin{tabular}{lccc}
	\hline
 		Methods & MPJPE~$\downarrow$ & PA-MPJPE~$\downarrow$ & MRPE~$\downarrow$ \\ \hline
 		SMPLify-X \cite{SMPL-X:2019} & 143.73 & 90.57 & 368.00\\
 		SPIN \cite{kolotouros2019spin} & 109.79 & 67.42 & 167.62\\
 		Baseline (\cite{kolotouros2019spin}+\cite{SMPL-X:2019}) & 82.92& 61.47 & 147.44 \\ \hline
 		Ours (w/o $L_{s}$, $L_{n}$) & 53.81 & 34.48 & 108.43 \\
 		Ours (w/o $L_{n}$) & 39.52 & 33.24 & 101.91 \\
 		\rowcolor{gray!10}
 		\textbf{Ours (full)} & \textbf{38.77} & \textbf{32.96} & \textbf{93.22} \\
  	    \hline
	\end{tabular}
	}
	\end{center}
	\caption{\textbf{Quantitative analysis.} `Baseline' uses \cite{kolotouros2019spin} as initialization and \cite{SMPL-X:2019} as the optimization method to fit the body model to 2D keypoints for each person separately. 
	}
	\label{tab:ablation}
\end{table}

%% file: tables/dataset.tex
\begin{table}[t]
	\begin{center}
	\resizebox{\columnwidth}{!}{
	\begin{tabular}{l*{3}{c}*{3}{c}}
	\hline
 		\multirow{2}[3]{*}{Datasets} & \multirow{2}[3]{*}{Frames(k)} & \multirow{2}[3]{*}{Subjects} & \multicolumn{3}{c}{Annotations} \\
 		\cmidrule(lr){4-6}
 		& & & 3D & 2D & Internet \\ \hline
 		InstaVariety \cite{humanMotionKanazawa19} & 2100 & 28272 & & & \cmark \\
 		Penn Action \cite{zhang2013actemes} & 77 & 2326 & & \cmark & \cmark \\
  		Human3.6M \cite{h36m_pami} & 581 & 11 &\cmark &\cmark & \\
 		3DPW \cite{vonMarcard2018} & 51 & 18 & \cmark & \cmark & \\
  		\rowcolor{gray!10}
 		Mirrored-Human & 1800 & 200+ & \cmark* & \cmark &\cmark   \\ 
		\hline
	\end{tabular}}
	\end{center}
	\caption{\textbf{Comparison of relevant datasets} in terms of the number of frames, subjects and annotation types. *Our Mirrored-Human dataset adopts our reconstructed 3D poses as pseudo ground-truth.}
	\label{tab:datasets}
\end{table}

%% file: src/05-1_learning.tex
\section{Learning with the Mirrored-Human dataset}
\subsection{Mirrored-Human}

Based on our framework, a large-scale Internet dataset can be built for the training of single-view tasks. 
The existing datasets~\cite{h36m_pami, mono-3dhp2017} lack the variety of both appearances and poses, making the training easy to overfit. For multi-person tasks, collecting data is more difficult . 
Therefore, previous methods exploit MuCo~\cite{mehta2018single}, a pseudo multi-person dataset composited from MPI-INF-3DHP~\cite{mono-3dhp2017} by masks, or JTA~\cite{fabbri2018learning}, a synthetic dataset. The gap between these datasets and the real scene may limit the performance of learning-based methods.

To alleviate the training data issue, we provide a large-scale Internet dataset named Mirrored-Human with our framework. Specifically, we collect a large number of videos from the Internet, in which we can see the person and the person’s mirror image. Actions cover dancing, fitness, mirror installation, swing practice, etc. Fig.~\ref{fig:dataset} demonstrates both the appearance and pose diversity of our dataset. 
Table~\ref{tab:datasets} shows a thorough comparison between our dataset and relevant datasets. 
Please refer to our supplementary material for more details of our dataset.

\subsection{Single-person mesh recovery}

\input{tables/sota}

For this task, we choose MeshNet~\cite{Moon_2020_ECCV_I2L-MeshNet}, a state-of-the-art method for single-view 3D pose estimation, for evaluation. Two datasets are used for evaluation. Human3.6M~\cite{h36m_pami} is an indoor benchmark with 3D annotations. 3DPW~\cite{vonMarcard2018} is an outdoor benchmark to test the generalization ability and only its defined test set is used. Following standard protocols, we report both MPJPE and PA-MPJPE. We also test the baseline method that uses the state-of-the-art optimization-based method SMPLify-X~\cite{SMPL-X:2019} to generate pseudo ground-truth to train the same network. Table~\ref{tab:sota_mesh} shows that with our dataset, the performance of MeshNet can be improved significantly, especially when tested on the 3DPW dataset without using training data from 3DPW. We also outperform the baseline, indicating that our framework is more accurate than \cite{SMPL-X:2019}.

\subsection{Multi-person 3D pose estimation.}
\begin{table}[t]
	\begin{center}
	\resizebox{\columnwidth}{!}{
	\begin{tabular}{clccc}
	\hline
 		& Methods & AP$_{root}^{25}\uparrow$ & PCK$_{rel}\uparrow$ & PCK$_{abs}\uparrow$\\ \hline
 		\multirow{7}{*}{TD} & LCRNet~\cite{rogez2017lcr} & - & 53.8 & - \\
 		& LCRNet++~\cite{rogez2019lcr} & - & 70.6 & - \\
 		& Dabral.~\cite{dabral2019multi} & - & 71.3 & - \\
 		& PandaNet~\cite{benzine2020pandanet} & - & 72.0 & -\\
 		& HMOR~\cite{li2020hmor} & - & 82.0 & \textbf{43.8} \\
        & Moon.~\cite{Moon_2019_ICCV_3DMPPE} & 31.0 & 81.8 & 31.5 \\
        \rowcolor{gray!10}
 		&Moon.~\cite{Moon_2019_ICCV_3DMPPE}+MiHu & \textbf{42.2} & \textbf{82.3} & 43.0 \\ \hline
 		\multirow{4}{*}{BU} & Mehta.~\cite{mehta2018single} & - & 65.0 & - \\
 		& Xnect~\cite{mehta2019xnect} & - & 70.4 & - \\
 		&SMAP~\cite{zhen2020smap} & 37.3 & 73.5 & 35.4 \\
 		\rowcolor{gray!10}
 		&SMAP~\cite{zhen2020smap}+MiHu & \textbf{42.3} & \textbf{74.1} & \textbf{38.0} \\ 
  	    \hline
	\end{tabular}
	}
	\end{center}
	\caption{Results on the MuPoTS-3D dataset. The numbers are calculated for all people. `MiHu' is our Mirrored-Human dataset. `TD' and `BU' mean `top-down' and `bottom-up', respectively.}
	\label{tab:sota_multi}
\end{table} 
For this task, previous methods fall into two categories. Top-down methods detect human first and then estimate keypoints with a single-person pose estimator. Bottom-up methods localize all keypoints in the image first and then group them into people. We choose the top-down method~\cite{Moon_2019_ICCV_3DMPPE} and the bottom-up method~\cite{zhen2020smap} for evaluation.

The MuPoTS-3D~\cite{mehta2018single} dataset is used. Following previous methods~\cite{Moon_2019_ICCV_3DMPPE,zhen2020smap}, AP$_{root}^{25}$, PCK$_{rel}$ and PCK$_{abs}$ are measured. AP$_{root}^{25}$ is the average precision of 3D human root location, which treats the prediction as correct if it lies within 25cm from the ground-truth. PCK$_{rel}$ is the percentage of correct keypoints after root alignment. A keypoint is correct if the distance between the prediction and the ground-truth is smaller than 15cm. PCK$_{abs}$ has almost the same definition as PCK$_{rel}$, but without the root alignment it measures the absolute pose accuracy. Note that AP is calculated only for the root and PCK is for all keypoints.

It can be observed from Table~\ref{tab:sota_multi} that with our dataset, AP$_{root}^{25}$ and PCK$_{abs}$ are improved significantly compared with \cite{Moon_2019_ICCV_3DMPPE}. For bottom-up methods, we also improve the performance of \cite{zhen2020smap} apparently.

%% file: tables/sota.tex
\begin{table}[t]
	\begin{center}
	\resizebox{\columnwidth}{!}{
	\begin{tabular}{l*{2}{c}*{2}{c}}
	\hline
 		\multirow{2}[3]{*}{Methods} & \multicolumn{2}{c}{3DPW} & \multicolumn{2}{c}{Human3.6M} \\ 
 		\cmidrule(lr){2-3} \cmidrule(lr){4-5}
 		& MPJPE~$\downarrow$ & PA-MPJPE~$\downarrow$ & MPJPE~$\downarrow$ & PA-MPJPE~$\downarrow$ \\ \hline
  		HMR~\cite{hmrKanazawa17} & - & 81.3 & 88.0 & 56.8 \\
  	    HMMR~\cite{humanMotionKanazawa19} & - & 72.6 & - & 56.9 \\
  	    Arnab.~\cite{Arnab_CVPR_2019} & - & 72.2 & 77.8 & 54.3\\
  	    CMR~\cite{kolotouros2019cmr} & - & 70.2 & - & 50.1 \\
  	    SPIN~\cite{kolotouros2019spin} & 98.2* & 59.2 & 62.3* & 41.1 \\
  	    MeshNet~\cite{Moon_2020_ECCV_I2L-MeshNet} & 93.2 & 58.6 & 55.7 & 41.7 \\
  	    Baseline & 90.0 & 57.5 & 54.7 & 41.7\\
  	     \rowcolor{gray!10}
  	    \cite{Moon_2020_ECCV_I2L-MeshNet}+MiHu & \textbf{85.1} & \textbf{54.8} & \textbf{53.6} & \textbf{41.0} \\
  	    \hline
	\end{tabular}
	}
	\end{center}
	\caption{Results on 3DPW and Human3.6M datasets. `MiHu' is our Mirrored-Human dataset. `Baseline' means training MeshNet~\cite{Moon_2020_ECCV_I2L-MeshNet} on Mirrored-Human with 3D annotations given by SMPLify-X~\cite{SMPL-X:2019}. *MPJPE of \cite{kolotouros2019spin} is obtained by their released model.}
	\vspace{-2mm}
	\label{tab:sota_mesh}
\end{table}

%% file: src/06_conclusion.tex
\section{Conclusion}
In this paper, we present an optimization-based framework that leverages the mirror reflection to reconstruct accurate 3D human pose. We collect a large-scale Internet dataset named Mirrored-Human with our reconstructed 3D poses as pseudo ground-truth and show that training on this dataset can enhance the performance of existing 3D human pose estimators. Our work opens many new directions for future research. We plan to extend the method to handle multiple mirrors, multiple people, multiple frames and more detailed reconstruction of shape and appearance.
\paragraph{Acknowledgement: }The authors would like to acknowledge the support from the National Key Research and Development Program of China (No. 2020AAA0108901) and NSFC (No. 61806176).

\input{figures/qual}
\input{figures/comparison}

%% file: figures/qual.tex
\begin{figure*}[ht]
\begin{center}
    \includegraphics[width=\textwidth, trim=10 330 100 10]{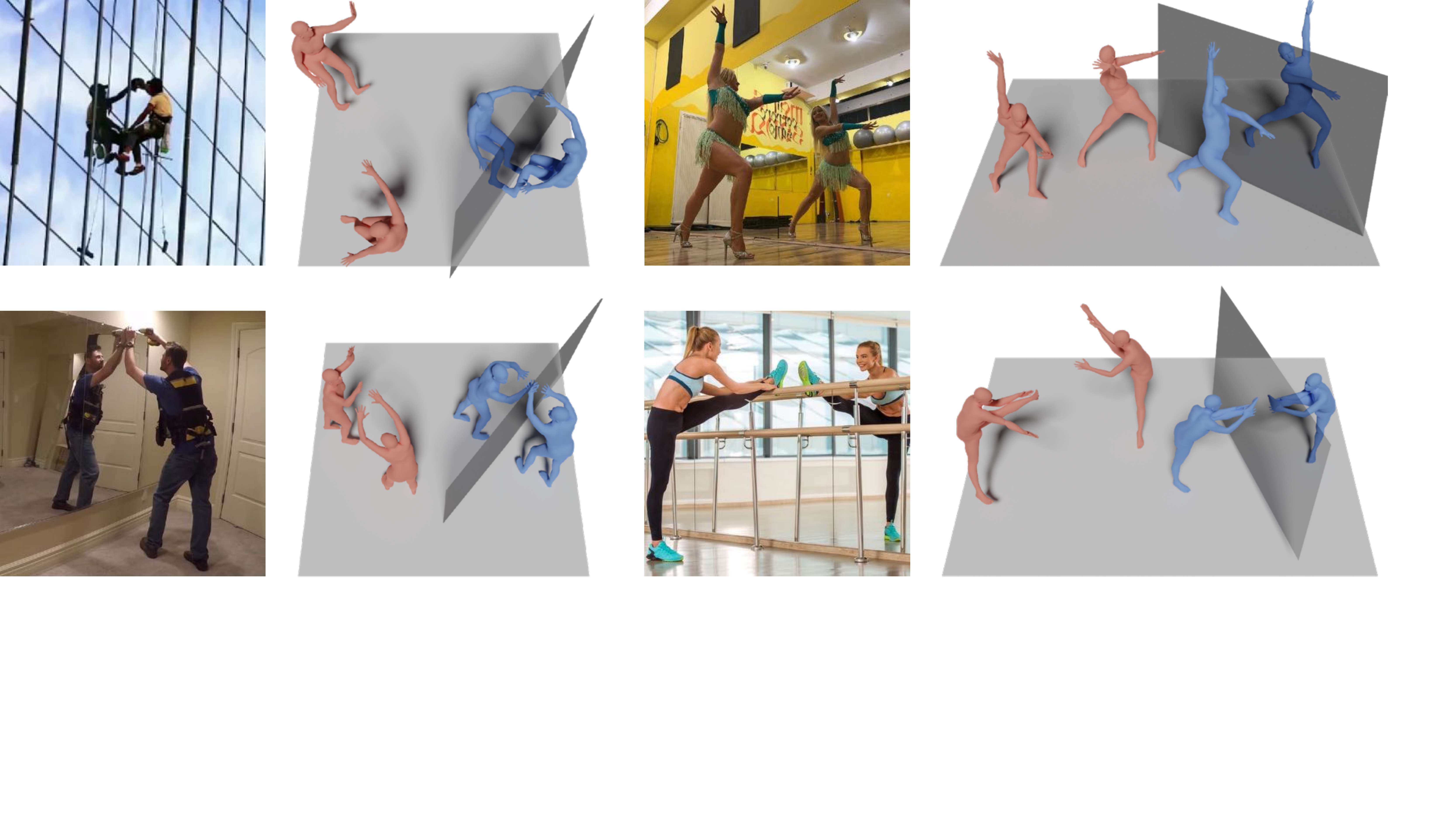}
\end{center}
\vspace{-2mm}
\footnotesize{\hspace{15mm}\text{Input}\hspace{15mm}\text{SMPLify-X\cite{SMPL-X:2019} (left) and ours (right)}\hspace{15mm}\text{Input}\hspace{27mm}\text{SMPLify-X\cite{SMPL-X:2019} (left) and ours (right)}\hspace{3mm}}
\vspace{1mm}
\caption{\textbf{Reconstruction results of Internet images.} Given the same 2D keypoints, we compare our method (blue) with SMPLify-X~\cite{SMPL-X:2019} (red). Our method simultaneously reconstructs precise poses, global locations and mirror geometry, while SMPLify-X~\cite{SMPL-X:2019} produces inconsistent results due to the depth ambiguity. 
}
\label{fig:results}
\end{figure*}

%% file: figures/comparison.tex
\begin{figure*}[!ht]
	\begin{center}
    	\includegraphics[width=0.48\linewidth, trim=0 30 270 50,clip]{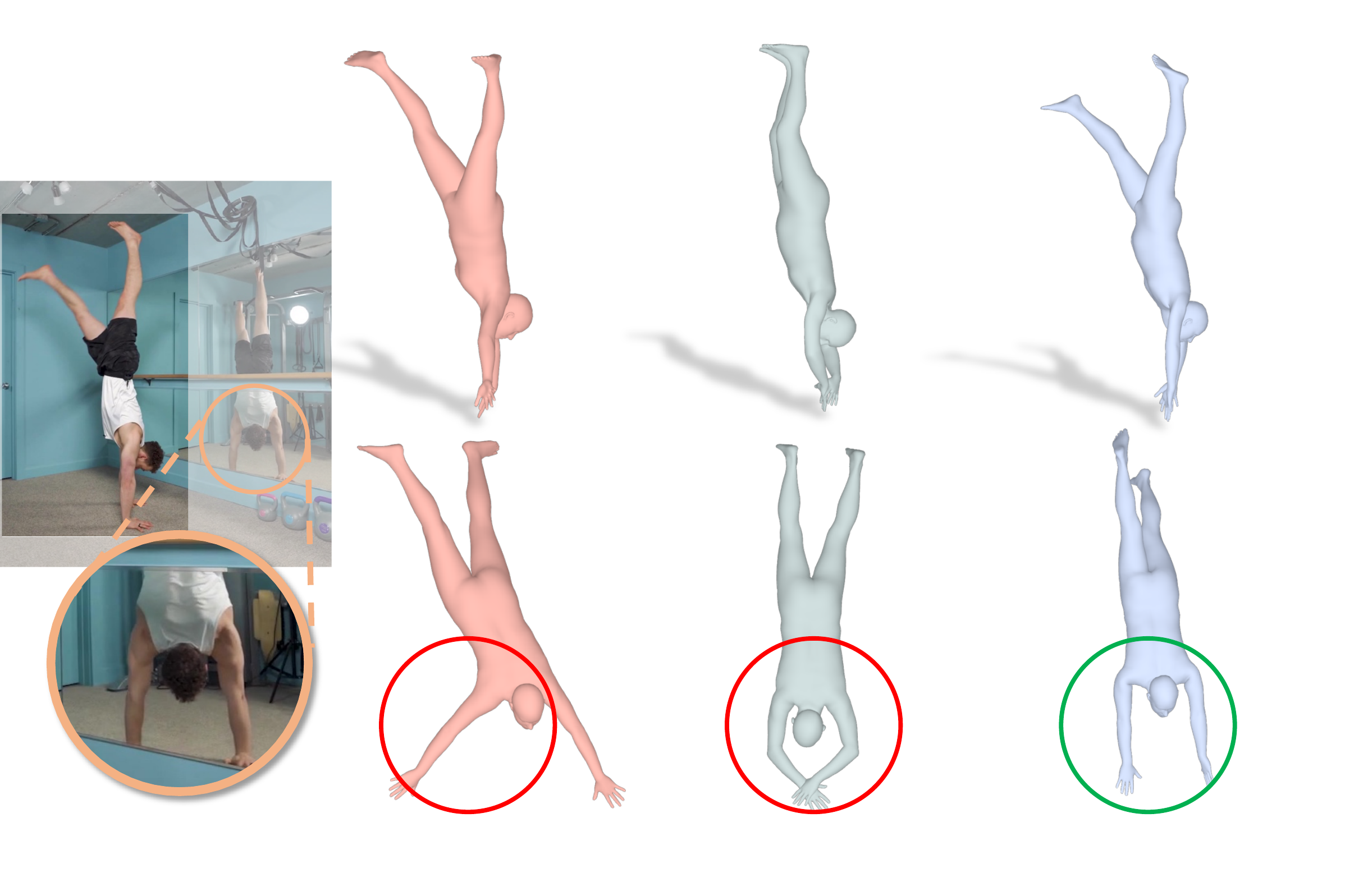}
    	\includegraphics[width=0.48\linewidth, trim=0 30 270 50,clip]{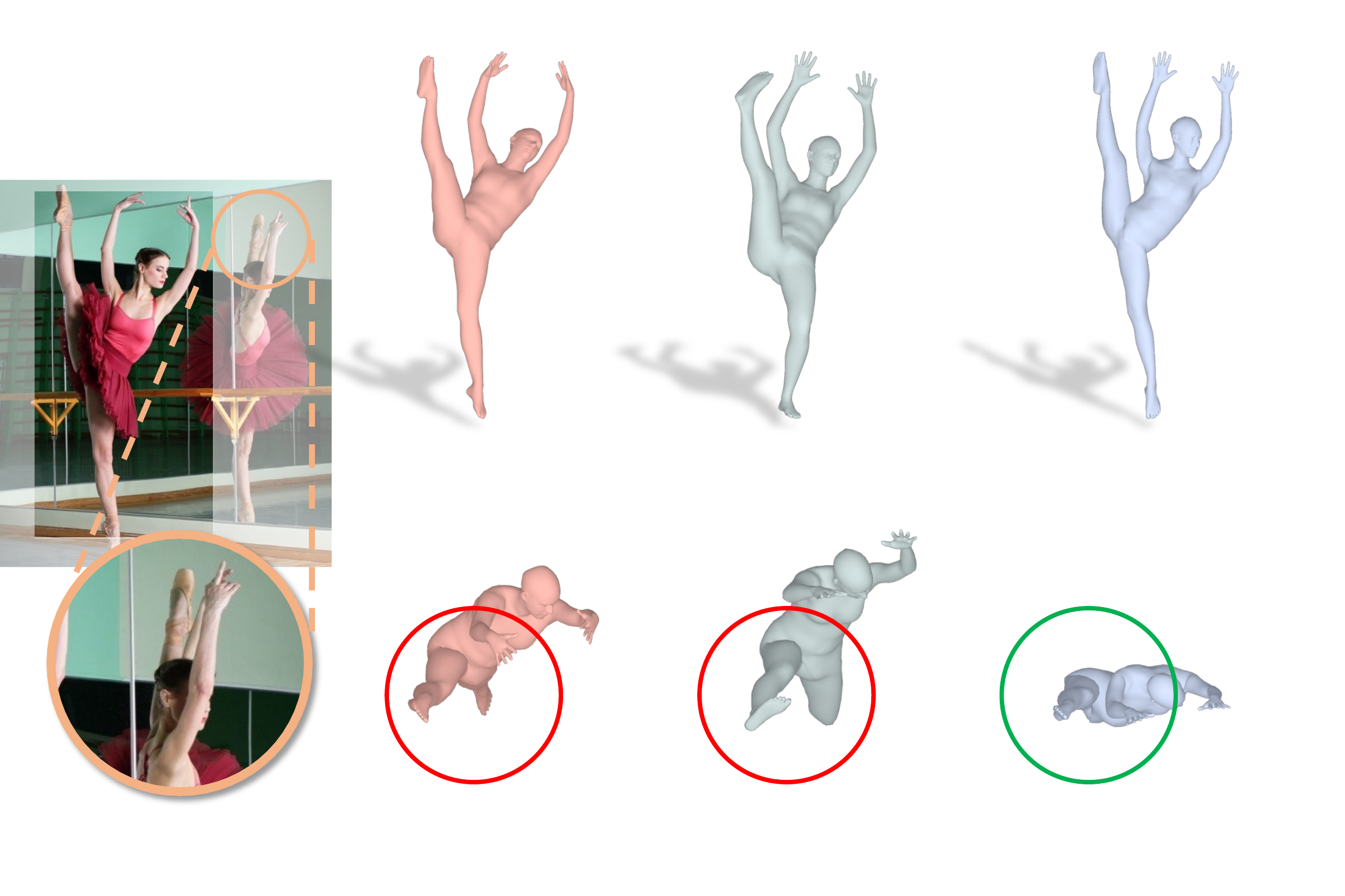}
    	\includegraphics[width=0.48\linewidth, trim=0 80 270 80,clip]{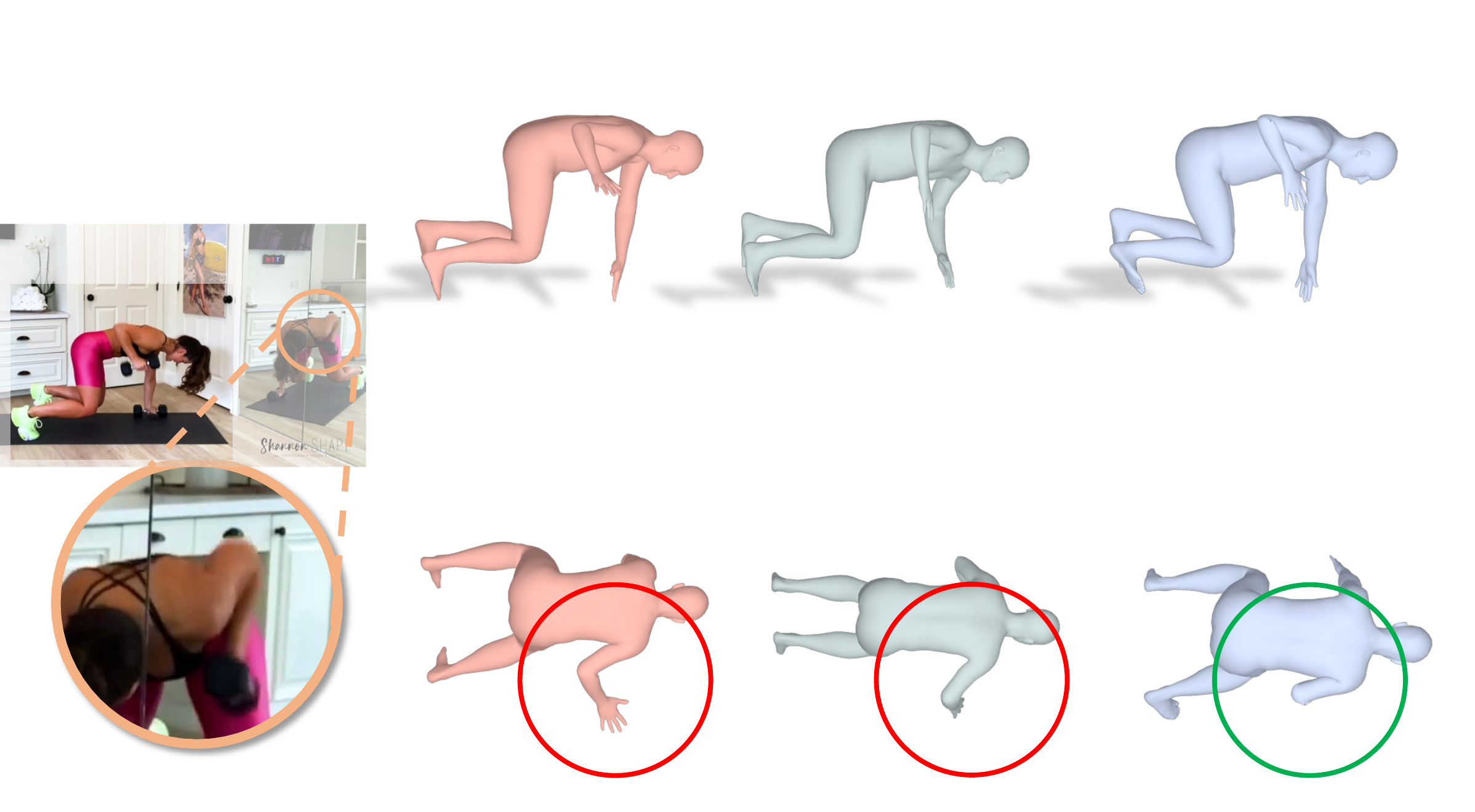}
    	\includegraphics[width=0.48\linewidth, trim=0 80 270 80,clip]{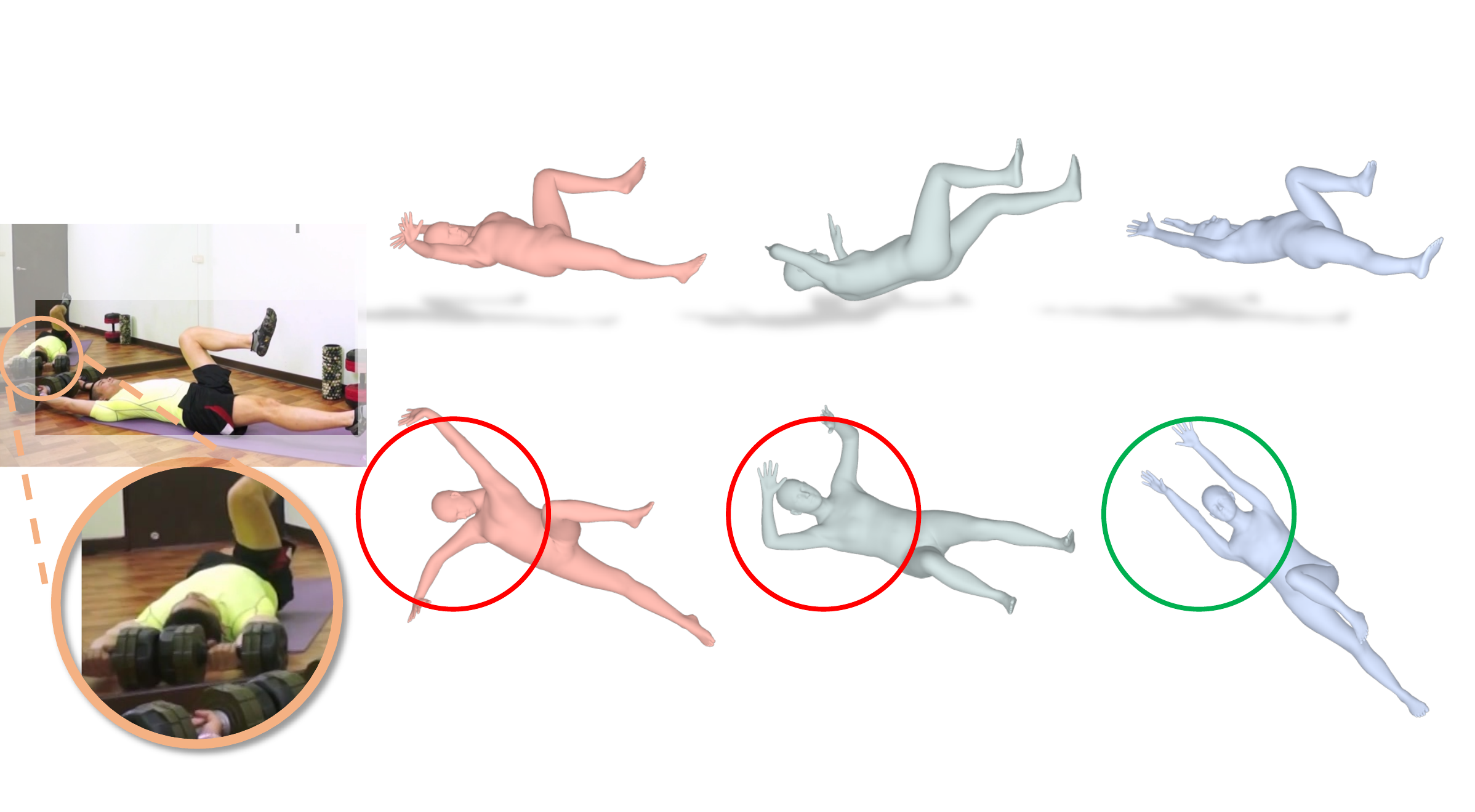}
	\end{center}
	\vspace{-5mm}
	\footnotesize{\hspace{6mm}\text{Input Image}\hspace{6mm}\text{SMPLify-X\cite{SMPL-X:2019}}\hspace{6mm}\text{SPIN\cite{kolotouros2019spin}}\hspace{13mm}\text{Ours}\hspace{12mm}\text{Input Image}\hspace{6mm}\text{SMPLify-X\cite{SMPL-X:2019}}\hspace{6mm}\text{SPIN\cite{kolotouros2019spin}}\hspace{13mm}\text{Ours}}
	\vspace{2mm}
	\caption{\textbf{Qualitative comparison of 3D pose estimation.} For each image, we show the results of the person estimated by the optimization-based method SMPLify-X~\cite{SMPL-X:2019} (red), the CNN-based method SPIN~\cite{kolotouros2019spin} (green) and our method (blue). For each block, top row shows the predicted mesh from the camera view and the bottom row shows another view. The circles emphasize some representative differences among three methods. The single-view optimization-based method produces inaccurate 3D poses despite smaller reprojection error. Our method produces more accurate 3D outputs given the same input.}
	\label{fig:comparisonpose}
	\vspace{-1cm}
\end{figure*}

%% file: cvpr.bbl
\begin{thebibliography}{10}\itemsep=-1pt

\bibitem{akay2014}
A. {Akay} and Y.~S. {Akgul}.
\newblock 3d reconstruction with mirrors and rgb-d cameras.
\newblock In {\em VISAPP}, 2014.

\bibitem{anguelov2005scape}
Dragomir Anguelov, Praveen Srinivasan, Daphne Koller, Sebastian Thrun, Jim
  Rodgers, and James Davis.
\newblock Scape: shape completion and animation of people.
\newblock In {\em SIGGRAPH}. 2005.

\bibitem{Arnab_CVPR_2019}
Anurag* Arnab, Carl* Doersch, and Andrew Zisserman.
\newblock Exploiting temporal context for 3d human pose estimation in the wild.
\newblock In {\em CVPR}, 2019.

\bibitem{benzine2020pandanet}
Abdallah Benzine, Florian Chabot, Bertrand Luvison, Quoc~Cuong Pham, and
  Catherine Achard.
\newblock Pandanet: Anchor-based single-shot multi-person 3d pose estimation.
\newblock In {\em CVPR}, 2020.

\bibitem{Bogo:ECCV:2016}
Federica Bogo, Angjoo Kanazawa, Christoph Lassner, Peter Gehler, Javier Romero,
  and Michael~J. Black.
\newblock Keep it {SMPL}: Automatic estimation of {3D} human pose and shape
  from a single image.
\newblock In {\em ECCV}, 2016.

\bibitem{cao2017realtime}
Zhe Cao, Tomas Simon, Shih-En Wei, and Yaser Sheikh.
\newblock Realtime multi-person 2d pose estimation using part affinity fields.
\newblock In {\em CVPR}, 2017.

\bibitem{chen20173d}
Ching-Hang Chen and Deva Ramanan.
\newblock 3d human pose estimation= 2d pose estimation+ matching.
\newblock In {\em CVPR}, 2017.

\bibitem{chen2019unsupervised}
Ching-Hang Chen, Ambrish Tyagi, Amit Agrawal, Dylan Drover, Stefan Stojanov,
  and James~M Rehg.
\newblock Unsupervised 3d pose estimation with geometric self-supervision.
\newblock In {\em CVPR}, 2019.

\bibitem{Choi_2020_ECCV_Pose2Mesh}
Hongsuk Choi, Gyeongsik Moon, and Kyoung~Mu Lee.
\newblock Pose2mesh: Graph convolutional network for 3d human pose and mesh
  recovery from a 2d human pose.
\newblock In {\em ECCV}, 2020.

\bibitem{dabral2019multi}
Rishabh Dabral, Nitesh~B Gundavarapu, Rahul Mitra, Abhishek Sharma, Ganesh
  Ramakrishnan, and Arjun Jain.
\newblock Multi-person 3d human pose estimation from monocular images.
\newblock In {\em 3DV}, 2019.

\bibitem{dong2020motion}
Junting Dong, Qing Shuai, Yuanqing Zhang, Xian Liu, Xiaowei Zhou, and Hujun
  Bao.
\newblock Motion capture from internet videos.
\newblock In {\em ECCV}, 2020.

\bibitem{fabbri2020compressed}
Matteo Fabbri, Fabio Lanzi, Simone Calderara, Stefano Alletto, and Rita
  Cucchiara.
\newblock Compressed volumetric heatmaps for multi-person 3d pose estimation.
\newblock In {\em CVPR}, 2020.

\bibitem{fabbri2018learning}
Matteo Fabbri, Fabio Lanzi, Simone Calderara, Andrea Palazzi, Roberto Vezzani,
  and Rita Cucchiara.
\newblock Learning to detect and track visible and occluded body joints in a
  virtual world.
\newblock In {\em ECCV}, 2018.

\bibitem{fieraru3d}
M. {Fieraru}, M. {Zanfir}, E. {Oneata}, A.~I. {Popa}, V. {Olaru}, and C.
  {Sminchisescu}.
\newblock Three-dimensional reconstruction of human interactions.
\newblock In {\em CVPR}, 2020.

\bibitem{gluckman2001catadioptric}
Joshua Gluckman and Shree~K Nayar.
\newblock Catadioptric stereo using planar mirrors.
\newblock {\em IJCV}, 2001.

\bibitem{hartley2003multiple}
Richard Hartley and Andrew Zisserman.
\newblock {\em Multiple view geometry in computer vision}.
\newblock Cambridge university press, 2003.

\bibitem{PROX:2019}
Mohamed Hassan, Vasileios Choutas, Dimitrios Tzionas, and Michael~J. Black.
\newblock Resolving {3D} human pose ambiguities with {3D} scene constraints.
\newblock In {\em ICCV}, 2019.

\bibitem{Hu2005MultipleView3R}
B. Hu, C.~M. Brown, and R. Nelson.
\newblock Multiple-view 3-d reconstruction using a mirror.
\newblock 2005.

\bibitem{h36m_pami}
Catalin Ionescu, Dragos Papava, Vlad Olaru, and Cristian Sminchisescu.
\newblock Human3.6m: Large scale datasets and predictive methods for 3d human
  sensing in natural environments.
\newblock {\em IEEE TPAMI}, 2014.

\bibitem{jiang2009symmetric}
Nianjuan Jiang, Ping Tan, and Loong-Fah Cheong.
\newblock Symmetric architecture modeling with a single image.
\newblock In {\em ACM SIGGRAPH Asia}. 2009.

\bibitem{Joo_2017_TPAMI}
Hanbyul Joo, Tomas Simon, Xulong Li, Hao Liu, Lei Tan, Lin Gui, Sean Banerjee,
  Timothy~Scott Godisart, Bart Nabbe, Iain Matthews, Takeo Kanade, Shohei
  Nobuhara, and Yaser Sheikh.
\newblock Panoptic studio: A massively multiview system for social interaction
  capture.
\newblock {\em IEEE TPAMI}, 2017.

\bibitem{hmrKanazawa17}
Angjoo Kanazawa, Michael~J. Black, David~W. Jacobs, and Jitendra Malik.
\newblock End-to-end recovery of human shape and pose.
\newblock In {\em CVPR}, 2018.

\bibitem{humanMotionKanazawa19}
Angjoo Kanazawa, Jason~Y. Zhang, Panna Felsen, and Jitendra Malik.
\newblock Learning 3d human dynamics from video.
\newblock In {\em CVPR}, 2019.

\bibitem{kocabas2019vibe}
Muhammed Kocabas, Nikos Athanasiou, and Michael~J. Black.
\newblock Vibe: Video inference for human body pose and shape estimation.
\newblock In {\em CVPR}, 2020.

\bibitem{kolotouros2019spin}
Nikos Kolotouros, Georgios Pavlakos, Michael~J Black, and Kostas Daniilidis.
\newblock Learning to reconstruct 3d human pose and shape via model-fitting in
  the loop.
\newblock In {\em ICCV}, 2019.

\bibitem{kolotouros2019cmr}
Nikos Kolotouros, Georgios Pavlakos, and Kostas Daniilidis.
\newblock Convolutional mesh regression for single-image human shape
  reconstruction.
\newblock In {\em CVPR}, 2019.

\bibitem{lanman2009surround}
Douglas Lanman, Daniel Crispell, and Gabriel Taubin.
\newblock Surround structured lighting: 3-d scanning with orthographic
  illumination.
\newblock {\em CVIU}, 2009.

\bibitem{Lassner:UP:2017}
Christoph Lassner, Javier Romero, Martin Kiefel, Federica Bogo, Michael~J.
  Black, and Peter~V. Gehler.
\newblock Unite the people: Closing the loop between 3d and 2d human
  representations.
\newblock In {\em CVPR}, 2017.

\bibitem{li2020hmor}
Jiefeng Li, Can Wang, Wentao Liu, Chen Qian, and Cewu Lu.
\newblock Hmor: Hierarchical multi-person ordinal relations for monocular
  multi-person 3d pose estimation.
\newblock In {\em ECCV}, 2020.

\bibitem{lin2020hdnet}
Jiahao Lin and Gim~Hee Lee.
\newblock Hdnet: Human depth estimation for multi-person camera-space
  localization.
\newblock In {\em ECCV}, 2020.

\bibitem{SMPL:2015}
Matthew Loper, Naureen Mahmood, Javier Romero, Gerard Pons-Moll, and Michael~J.
  Black.
\newblock {SMPL}: A skinned multi-person linear model.
\newblock {\em ACM TOG}, 2015.

\bibitem{martinez_2017_3dbaseline}
Julieta Martinez, Rayat Hossain, Javier Romero, and James~J. Little.
\newblock A simple yet effective baseline for 3d human pose estimation.
\newblock In {\em ICCV}, 2017.

\bibitem{mono-3dhp2017}
Dushyant Mehta, Helge Rhodin, Dan Casas, Pascal Fua, Oleksandr Sotnychenko,
  Weipeng Xu, and Christian Theobalt.
\newblock Monocular 3d human pose estimation in the wild using improved cnn
  supervision.
\newblock In {\em 3DV}, 2017.

\bibitem{mehta2019xnect}
Dushyant Mehta, Oleksandr Sotnychenko, Franziska Mueller, Weipeng Xu, Mohamed
  Elgharib, Pascal Fua, Hans-Peter Seidel, Helge Rhodin, Gerard Pons-Moll, and
  Christian Theobalt.
\newblock Xnect: Real-time multi-person 3d human pose estimation with a single
  rgb camera.
\newblock {\em ACM TOG}, 2020.

\bibitem{mehta2018single}
Dushyant Mehta, Oleksandr Sotnychenko, Franziska Mueller, Weipeng Xu, Srinath
  Sridhar, Gerard Pons-Moll, and Christian Theobalt.
\newblock Single-shot multi-person 3d pose estimation from monocular rgb.
\newblock In {\em 3DV}, 2018.

\bibitem{Moon_2019_ICCV_3DMPPE}
Gyeongsik Moon, Juyong Chang, and Kyoung~Mu Lee.
\newblock Camera distance-aware top-down approach for 3d multi-person pose
  estimation from a single rgb image.
\newblock In {\em ICCV}, 2019.

\bibitem{Moon_2020_ECCV_I2L-MeshNet}
Gyeongsik Moon and Kyoung~Mu Lee.
\newblock I2l-meshnet: Image-to-lixel prediction network for accurate 3d human
  pose and mesh estimation from a single rgb image.
\newblock In {\em ECCV}, 2020.

\bibitem{nene1998}
S.~A. {Nene} and S.~K. {Nayar}.
\newblock Stereo with mirrors.
\newblock In {\em ICCV}, 1998.

\bibitem{nguyen20183d}
Trong-Nguyen Nguyen, Huu-Hung Huynh, and Jean Meunier.
\newblock 3d reconstruction with time-of-flight depth camera and multiple
  mirrors.
\newblock {\em IEEE Access}, 2018.

\bibitem{omran2018nbf}
Mohamed Omran, Christoph Lassner, Gerard Pons-Moll, Peter~V. Gehler, and Bernt
  Schiele.
\newblock Neural body fitting: Unifying deep learning and model-based human
  pose and shape estimation.
\newblock In {\em 3DV}, 2018.

\bibitem{STAR:2020}
Ahmed A~A Osman, Timo Bolkart, and Michael~J. Black.
\newblock {STAR}: A spare trained articulated human body regressor.
\newblock In {\em ECCV}, 2020.

\bibitem{SMPL-X:2019}
Georgios Pavlakos, Vasileios Choutas, Nima Ghorbani, Timo Bolkart, Ahmed A.~A.
  Osman, Dimitrios Tzionas, and Michael~J. Black.
\newblock Expressive body capture: 3d hands, face, and body from a single
  image.
\newblock In {\em CVPR}, 2019.

\bibitem{pavlakos2018ordinal}
Georgios Pavlakos, Xiaowei Zhou, and Kostas Daniilidis.
\newblock Ordinal depth supervision for 3d human pose estimation.
\newblock In {\em CVPR}, 2018.

\bibitem{peng2021neural}
Sida Peng, Yuanqing Zhang, Yinghao Xu, Qianqian Wang, Qing Shuai, Hujun Bao,
  and Xiaowei Zhou.
\newblock Neural body: Implicit neural representations with structured latent
  codes for novel view synthesis of dynamic humans.
\newblock In {\em CVPR}, 2021.

\bibitem{rodrigues2010}
Rui Rodrigues, Joao~P Barreto, and Urbano Nunes.
\newblock Camera pose estimation using images of planar mirror reflections.
\newblock In {\em ECCV}, 2010.

\bibitem{rogez2017lcr}
Gregory Rogez, Philippe Weinzaepfel, and Cordelia Schmid.
\newblock Lcr-net: Localization-classification-regression for human pose.
\newblock In {\em CVPR}, 2017.

\bibitem{rogez2019lcr}
Gregory Rogez, Philippe Weinzaepfel, and Cordelia Schmid.
\newblock Lcr-net++: Multi-person 2d and 3d pose detection in natural images.
\newblock {\em IEEE TPAMI}, 2019.

\bibitem{Sigal:IJCV:10b}
L. Sigal, A. Balan, and M.~J. Black.
\newblock {HumanEva}: Synchronized video and motion capture dataset and
  baseline algorithm for evaluation of articulated human motion.
\newblock {\em IJCV}, 2010.

\bibitem{sinha2012detecting}
Sudipta~N Sinha, Krishnan Ramnath, and Richard Szeliski.
\newblock Detecting and reconstructing 3d mirror symmetric objects.
\newblock In {\em ECCV}, 2012.

\bibitem{sun2019deep}
Ke Sun, Bin Xiao, Dong Liu, and Jingdong Wang.
\newblock Deep high-resolution representation learning for human pose
  estimation.
\newblock In {\em CVPR}, 2019.

\bibitem{sun2017compositional}
Xiao Sun, Jiaxiang Shang, Shuang Liang, and Yichen Wei.
\newblock Compositional human pose regression.
\newblock In {\em ICCV}, 2017.

\bibitem{sun2018integral}
Xiao Sun, Bin Xiao, Fangyin Wei, Shuang Liang, and Yichen Wei.
\newblock Integral human pose regression.
\newblock In {\em ECCV}, 2018.

\bibitem{tahara2015interference}
Tomu Tahara, Ryo Kawahara, Shohei Nobuhara, and Takashi Matsuyama.
\newblock Interference-free epipole-centered structured light pattern for
  mirror-based multi-view active stereo.
\newblock In {\em 3DV}, 2015.

\bibitem{takahashi2012}
K. {Takahashi}, S. {Nobuhara}, and T. {Matsuyama}.
\newblock A new mirror-based extrinsic camera calibration using an
  orthogonality constraint.
\newblock In {\em CVPR}, 2012.

\bibitem{tripathi2020posenet3d}
Shashank Tripathi, Siddhant Ranade, Ambrish Tyagi, and Amit Agrawal.
\newblock Posenet3d: Learning temporally consistent 3d human pose via knowledge
  distillation.
\newblock In {\em 3DV}, 2020.

\bibitem{vonMarcard2018}
Timo von Marcard, Roberto Henschel, Michael Black, Bodo Rosenhahn, and Gerard
  Pons-Moll.
\newblock Recovering accurate 3d human pose in the wild using imus and a moving
  camera.
\newblock In {\em ECCV}, 2018.

\bibitem{wandt2019repnet}
Bastian Wandt and Bodo Rosenhahn.
\newblock Repnet: Weakly supervised training of an adversarial reprojection
  network for 3d human pose estimation.
\newblock In {\em CVPR}, 2019.

\bibitem{Wu_2020_CVPR}
Shangzhe Wu, Christian Rupprecht, and Andrea Vedaldi.
\newblock Unsupervised learning of probably symmetric deformable 3d objects
  from images in the wild.
\newblock In {\em CVPR}, 2020.

\bibitem{xiang2019monocular}
Donglai Xiang, Hanbyul Joo, and Yaser Sheikh.
\newblock Monocular total capture: Posing face, body, and hands in the wild.
\newblock In {\em CVPR}, 2019.

\bibitem{xu2020ghum}
Hongyi Xu, Eduard~Gabriel Bazavan, Andrei Zanfir, William~T Freeman, Rahul
  Sukthankar, and Cristian Sminchisescu.
\newblock Ghum \& ghuml: Generative 3d human shape and articulated pose models.
\newblock In {\em CVPR}, 2020.

\bibitem{xu2019denserac}
Yuanlu Xu, Song-Chun Zhu, and Tony Tung.
\newblock Denserac: Joint 3d pose and shape estimation by dense
  render-and-compare.
\newblock In {\em ICCV}, 2019.

\bibitem{ying2012self}
Xianghua Ying, Kun Peng, Yongbo Hou, Sheng Guan, Jing Kong, and Hongbin Zha.
\newblock Self-calibration of catadioptric camera with two planar mirrors from
  silhouettes.
\newblock {\em IEEE TPAMI}, 2012.

\bibitem{zanfir2018monocular}
Andrei Zanfir, Elisabeta Marinoiu, and Cristian Sminchisescu.
\newblock Monocular 3d pose and shape estimation of multiple people in natural
  scenes-the importance of multiple scene constraints.
\newblock In {\em CVPR}, 2018.

\bibitem{zeng20203d}
Wang Zeng, Wanli Ouyang, Ping Luo, Wentao Liu, and Xiaogang Wang.
\newblock 3d human mesh regression with dense correspondence.
\newblock In {\em CVPR}, 2020.

\bibitem{zhang2013actemes}
Weiyu Zhang, Menglong Zhu, and Konstantinos~G Derpanis.
\newblock From actemes to action: A strongly-supervised representation for
  detailed action understanding.
\newblock In {\em ICCV}, 2013.

\bibitem{zhen2020smap}
Jianan Zhen, Qi Fang, Jiaming Sun, Wentao Liu, Wei Jiang, Hujun Bao, and
  Xiaowei Zhou.
\newblock Smap: Single-shot multi-person absolute 3d pose estimation.
\newblock In {\em ECCV}, 2020.

\bibitem{zhou2017towards}
Xingyi Zhou, Qixing Huang, Xiao Sun, Xiangyang Xue, and Yichen Wei.
\newblock Towards 3d human pose estimation in the wild: a weakly-supervised
  approach.
\newblock In {\em ICCV}, 2017.

\end{thebibliography}
